# Annotation automatique des connaissances spatiales en arabe

*Rita Hijazi, Amani Sabra, Moustafa Al-Hajj*

Université Libanaise, Centre des Sciences du Langage et de la Communication, Faculté des Lettres, Tayouneh, Beyrouth, Liban

* rita.hijazi@hotmail.com, ** amani.sabra@hotmail.fr, ***moustafa.alhajj@ul.edu.lb

**Résumé :**

Dans cet article, nous introduisons une approche à base des règles pour l'annotation automatique des expressions de la localisation et de la direction en arabe. La visée de cet article est double : il s'agit premièrement de définir les structures linguistiques qui expriment la localisation et la direction en arabe, et deuxièmement d'envisager le problème de l'annotation automatique des segments textuels conformes à ces structures linguistiques. La méthodologie adoptée dans cette étude est analytique descriptive. En s'appuyant sur NOOJ, un outil pour l'implémentation des transducteurs à états finis, nous présentons un système d'annotation des connaissances spatiales en arabe, en se basant sur une analyse linguistique profonde et sur une carte sémantique du domaine de la spatialité. Les résultats obtenus à partir de l'application sur un roman spécifique ont révélé que la méthode préconisée semble être adéquate et affiche des résultats de reconnaissance très encourageants. En conclusion, NOOJ nous a permis l'écriture des règles linguistiques pour l'annotation automatique dans un texte en arabe des expressions de la localisation spatiale et de la direction.

**Abstract :**

In this paper, we introduce a rule-based approach to annotate Locative and Directional Expressions in Arabic natural language text. The annotation is based on a constructed semantic map of the spatiality domain. Challenges are twofold: first, we need to study how locative and directional expressions are expressed linguistically in these texts; and second, we need to automatically annotate the relevant textual segments accordingly. The research method we will use in this article is analytic-descriptive. We will validate this approach on specific novel rich with these expressions and show that it has very promising results. We will be using NOOJ as a software tool to implement finite-state transducers to annotate linguistic elements according to Locative and Directional Expressions. In conclusion, NOOJ allowed us to write linguistic rules for the automatic annotation in Arabic text of Locative and Directional Expressions.

**Mots-clés :** Extraction d'information, Expressions de localisation en arabe.

## 1. Introduction

L'arabe exprime des informations locatives et positionnelles très riches et intéressantes. Le travail se situe à la croisée de deux grandes perspectives :

(1) une perspective linguistique, d'une part, et (2) la perspective informatique, d'autre part.

Sur le plan linguistique, l'objectif initial de ce travail est de présenter un essai de classification générale des prépositions de localisation et de la direction en prenant en compte leurs propriétés sémantiques.

Le deuxième objectif, sur le plan informatique, concernait l'écriture de règles informatiques, implémentables dans la plateforme NooJ, permettant d'extraire automatiquement une expression de localisation dans un corpus arabe.

De nombreux travaux se réclament des grammaires cognitives de Jackendoff (1991, 1996), Herskovits (1986, 1997), Vandeloise (1986, 1992, 1999) et Talmy (2000), (pour ne citer que quelques références), qui portent sur la cognition de l'espace et sur la sémantique des prépositions spatiales.

Notre approche se base sur des études faites sur d'autres langues. Dans son étude, Kopecka, présente la typologie de l'expression de la localisation et du déplacement en français et en polonais dont le but était d'explorer l'expression de ces domaines sémantiques dans les deux langues dans une perspective typologique afin d'évaluer son impact sur l'élaboration linguistique de l'information spatiale (Kopecka, 2004).

En arabe, Mubarak, dans son livre fournit une analyse détaillée des effets de sens de plusieurs prépositions, sans se limiter à leurs emplois spatiaux. Il s'intéresse à étudier la sémantique et l'usage de chaque préposition (Mubarak, 1988).

L'intérêt particulier du choix de la langue arabe repose sur la rareté des travaux élaborés dans cette langue qui traitent ce sujet. En addition, dans notre étude, nous voulons montrer que l'ingénierie des langues peut exécuter les phénomènes de localisation et de direction spatiale. De plus, montrer que l'interaction avec les données décrivant des entités spatiales est possible afin de les enrichir d'une façon semi-automatique tout en se basant sur des outils informatiques déjà développés. Ce travail est l'un de nos travaux sur l'annotation sémantique de textes en langue arabes (Alhajj et Mourad, 2015), (Alhajj et Sabra, 2018).

Dans cet article nous essayons de présenter un système d'annotation pour l'arabe qui, après avoir effectué une analyse linguistique, annote les expressions de localisation. L'annotation se base sur une carte sémantique du domaine de la spatialité.

Nous commençons par présenter dans la section 2 le corpus de travail. La section 3 présente une description des étapes de l'analyse linguistique, une étude du système des prépositions de localisation en arabe. La section 4 décrit l'organisation du domaine spatial sous forme d'une carte sémantique. La section 5 sera réservée au processus d'annotation automatique qui se base sur les résultats de l'analyse linguistique et sur la carte, sous la forme des règles spécifiques au domaine traité : nous présentons le traitement qui consiste à annoter les informations recueillies, au niveau du corpus, accompagné par une évaluation du système ainsi qu'une analyse des erreurs. Nous clôturons par une conclusion.

## 2. Le corpus

Notre étude est appuyée sur les données collectées à partir d'un roman en arabe intitulé غريقة بحيرة موريه (*La naufragée du lac Murray*) écrit par Antoine Douaihy, en 2015, qui résume une histoire d'amour étrange, décrivant l'endroit où deux cultures se convergent et se divergent ; dans lesquels le narrateur s'immerge dans les mystères de l'âme humaine et tente de pénétrer le fond des deux univers.

Ce corpus apparaît comme caractéristique du texte narratif, du genre littéraire. Cela revient à la fréquence d'utilisation des prépositions, les adverbes et les déictiques spatiaux qui sont plutôt caractéristiques de la narration, du récit. Les exemples tirés sont traduits en français afin de faciliter leur compréhension pour la communauté francophone.

Notre but sera de collecter les occurrences des expressions de localisation dans le roman, les classer dans des listes semi-exhaustives et essayer d'étudier les ambiguïtés à travers les organisateurs collectés dans les divers contextes.

Le corpus a été divisé en deux parties : un corpus de travail qui constitue 75% de l'ensemble du corpus, il a permis d'identifier les ressources lexicales complémentaires nécessaires et de construire les listes et les patrons. La partie restante du corpus originale a constituée le corpus d'évaluation (25%) sur lequel la méthode d'annotation automatique a été évaluée.

## 3. Analyse linguistique

Le traitement automatique de l'Arabe est considéré comme difficile à appréhender, surtout les prépositions, qui sont abondamment utilisées dans l'expression d'autres domaines, notamment la temporalité.

L'analyse linguistique est nécessaire pour assurer une annotation automatique pertinente d'informations.

On peut définir deux types de relations entre le site et la cible (Aurnague, Vieu et Borillo 1997, Borillo 1998, Aurnague 2004, Costăchescu 2008a) :

1- **La relation topologique**, aussi appelée relation de «Localisation Interne». Si la cible est localisée dans une partie spatiale ayant une coïncidence avec le site, avec support ou comme inclusion, on parle d'une relation topologique.

2. **La relation projective** appelée aussi relation de «Localisation Externe». Si on peut distinguer la position de la cible par rapport à un site qui lui est extérieur, on parle d'une relation projective (Costăchescu, 2008a).

Les relations topologiques peuvent être subdivisées en deux catégories : la relation de support et la relation d'inclusion. Les relations projectives, ou de localisation externe, peuvent être classifiées en deux grandes catégories : les relations projectives de distance et les relations projectives orientationnelles. (Costăchescu, 2008b)

Les relations orientationnelles situent la cible sur l'un des trois axes : latéral, vertical ou frontal.

## 3.1. Etude du système des prépositions en arabe

### 3.1.1. Etude sémantique de la préposition « على » (sur)

جلسَت المرأة **على** المقعد. ( p.16)

*La femme est assise **sur** le banc.*

1) اتجهنا لورا وأنا سيرًا على القدمين **في محاذاة** نهر السين من جزيرة سان لويس إلى "القصر الكبير". (p.92)

*Nous nous sommes dirigées, Laura et moi, à pieds, **côtoyant** La Seine, allant de l'île Saint Louis jusqu'à le Grand-Palais.*

2) كما كنت أمضى فترة الانتظار **في** مقهى لوتيسيا المجاور، المطلّ على ساحة مونج، حيث اعتدتُ الكتابة [...] (p.10)

*Je passais de même la période d'attente **dans** le café voisin, Leuticia, donnant sur l'arène de Monge où j'avais l'habitude d'écrire.*

La préposition « على » (sur) marque:
  a. la position de la cible et du site par rapport à l'axe vertical, mais également le contact entre les entités (énoncé 1).

  b. une relation de support représentée par des nominaux qui désignent des parties du corps, une relation d'ingrédience (partie/tout) entre les entités (énoncé 2)

  c. la direction d'un regard (énoncé 3)

Il existe en arabe une norme qui distingue les emplois de « فوق» (*au-dessus*) et « على » (*sur*) :
"على" implique un contact entre les deux entités locatives, or, « فوق » ne l'impose pas. Appliquons le test de substitution sur l'exemple suivant:

3) كنت أسير **فوق** طبقة من الثلج.

*Je marchais **au-dessus** d'une couche de neige.*

En effet, nous pouvons indifféremment dire :

كنت أسير **على** طبقة من الثلج.

*Je marchais **sur** une couche de neige.*

Dans l'énoncé (5) ci-dessus, on ne peut pas remplacer la préposition فوق par على. Les deux entités « روزا » (*Rosa*) et l'auteur ne sont pas en contact, il s'agit de raconter que la famille habite dans le bâtiment juste après. Il ne s'agit pas ici non plus d'un cas où deux objets sont l'un au-dessus de l'autre, mais plutôt d'un nouvel exemple où ce sont les limites de la cible et du site (ici, bloc à trois dimensions) qui sont considérées.

4) كانت لي قبلها روزا المقيمة في الطبقة الخامسة **فوقي** وقد ربطتني بها مودّة خالصة. (p.24)

*Je connaissais Rosa qui vivait au cinquième étage, au-dessus de moi et un pur respect nous liait à elle.*

### 3.1.2. La sémantique des prépositions "عند"، "بين"، "داخل"، "في"، "حيث"

**a) L'intérieur d'un lieu**

5) الانتقال **داخل** بلاد البروفانس. (p.68)
*Le déplacement **dans** les pays des Provences.*

6) قالت إنها تحب التعارف إلى حي حديقة مونسوري **حيث** كنت أقيم.(p.16)
*Elle a dit qu'elle aimait voir le quartier **où** je vivais.*

7) جلست **في** المقهى وكتبت لها البطاقة البريدية الآتية (p.21)
*Je me suis assis **dans** le café et je lui ai écrit la carte postale suivante.*

8) كنا معا لميا وأنا **في** مقهى **عند** "شاطئ النخلتين"...(p.123)
*Nous étions ensemble, Lamia et moi, **dans** le café **à** la « plage des deux palmes ».*

Ces prépositions mettent en relation une entité cible et une entité site qui leur servent de contenant, ou qui l'inclut à l'intérieur de ses limites. Ces prépositions expriment une intériorité. Un point est dit intérieur à un lieu, s'il est inclus dans l'espace contenu par l'enveloppe d'un lieu. La même interprétation s'applique sur les locutions "في قلب" et "في صدر" dans les énoncés (10) et (11).

9) سرعان ما انتشر الخبر بأن سيارة مفخخة انفجرت قرب "الحديقة العامة" و "برج الساعة" العثماني **في قلب** المدينة، على بعد دقائق بالسيارة من هنا، موقعة عشرات القتلى والجرحى في صفوف المارة.( p.132)

*La nouvelle s'est rapidement répandue : une voiture piégée explosa près du jardin publique" et de la "Tour de l'Horloge" ottomane **au plein cœur** de la ville, à quelques minutes de conduit d'ici, lassant des dizaines de morts et de blessés parmi les piétons.*

10) كانت لوحة "درس الكتابة العجائبي" الماثلة **في صدر** الشقة تشيع في المكان جو ساحر، بلونها الأحمر الزهري والزيتي الغامق الغالبين[...] (p.141)

*La toile « la leçon de l'écriture miraculeuse » qui se trouvait **au centre de** l'appartement répandait dans l'endroit une atmosphère charmante, avec sa couleur rouge rosée et vert olive les plus marquantes.*

La préposition بين (entre) dans l'énoncé (12) indique l'inclusion mais dans le cas où l'entité est plurielle. On parle dans ce cas du concept de distribution.

11) ارتمت **بين** ذراعيّ.(p.17)
*Elle s'est endormie **entre** mes bras.*

Comme toutes les autres prépositions spatio-temporelles, وسط , عند , بين sont susceptibles d'usages spatiaux, (énoncés 9, 12 et 13) ; d'usages temporels (énoncés 14, 25 et 16) et d'usages abstraits (énoncés 17 et 18).

12) استغربت وجود قاربين حاطين **في وسط السهل**. (p.156)
*J'étais surpris de la présence de deux navires placés **au milieu** de la plaine.*

13) **بين** ساعة الغروب **ومنتصف** الليل (p.23)
***Entre** le coucher du soleil et **minuit**.*

14) **عند منتصف** الليل كنا مستلقيين جنبًا إلى جنب، ويدانا متشابكتان في سكينة هانئة. (p.95)

*A minuit*, nous étions couchés côte-à-côte, nos mains entrelacées dans une calme agréable.

15) هطل الثلج من جديد أيّامًا **أواسط** شباط. (p.183)

*La neige tomba de nouveau plusieurs jours en **mi**-février.*

16) ها أنا في طائرة العودة إلى باريس، بعد سنين طويلة من تلك الأحداث، سابح **في** الفضاء بين عالمين. (p.138)

*Me voici dans l'avion de retour de Paris, après beaucoup d'années de ces évènements-là, surfant **dans** l'espace entre deux mondes.*

17) أدرك بحدسي مسبقًا ثغرها ومتاهاتها **وسط** بحر من أشجان الروح ترتفع فوق الرغبة المستحيلة في تخطي الموت… (p.97)

*J'ai pu sentir intuitivement ses lacunes **au milieu** d'une mer des passions de l'âme s'élevant au-dessus du désir impossible du dépassement de la mort.*

**b) La notion de périphérie**

Un périphérique définit les relations entre cible et site sur la base de la partie périphérique du site, comme dans l'énoncé (19)

18) سرت طويلًا **على ضفة** اللواريه (p.151)

*J'ai marché longtemps **au bord** de la Loire.*

### 3.1.3. La sémantique de la préposition الباء

Il existe une relation d'ingrédience (partie/tout) entre l'entité qui correspond au syntagme nominal qui suit la préposition « الباء » et une entité représentée dans la phrase. Dans ce cas, on se rapproche d'une anaphore associative. Ainsi dans l'énoncé (20), "يد" (une main) est un ingrédient du corps humain. Une autre valeur existe entre les entités spatiales, la valeur de médiateur, un déplacement d'un lieu à un autre par un moyen de locomotion, un médiateur, c'est le cas de l'énoncé (21).

19) أحدنا (لورا وأنا) مُمسك بطمأنينة واستسلام عميقين **بيد** الآخر. (p.13)

*L'un de nous (Laura et moi) tient avec tranquillité et sécurité profonde **la main de** l'autre.*

20) غادر أخي رامي **بالباخرة،** لكنه عاد بالطائرة بعد نحو أسبوعين (p.63)

*Mon frère Rami a quitté en bateau mais il est revenu **en avion** après deux semaines environ.*

### 3.1.4. Description des locutions prépositionnelles يمين et يسار «*droite et gauche*»

Dans les énoncés (22) et (23), le site المدخل (*l'entrée*) présente une orientation **intrinsèque** [1]: cette entité se trouve à droite du صندوق البريد *(boite aux lettres)* de point de vue de l'observateur par rapport à sa position.

21) وعندما هممت بمغادرتها في اتجاه الغابة، وقع نظري صدفة على صندوق البريد القديم المعلّق **عن يمين** المدخل. (p.157)

*Lorsque j'étais sur la pointe de partir vers la forêt, mon regard se posa par hasard, sur l'ancienne boite aux lettres accrochée **à droite** de l'entrée.*

22) فرقة الأقارب التي التأمت بالمئات في كتلة واحدة **عن يمين** باب السيدة الملوكي تتقدمها عائلة رؤوف، تقابلها **عن يسار** الباب كتلة المعزين التي ضمت الاف الناس. (p.41)

*Les proches qui s'entrelaçaient par centaine en un seul bloc **à droite** de la porte Royale de la Dame, dirigée par la famille de Raouf, alors qu'**à gauche**, se trouvait le groupe des pleureurs qui comptait des milliers de personnes.*

En présence des noms يمين et يسار le site regagne son autonomie si les deux substantifs sont accompagnés par un adjectif possessif, souvent nécessaire pour éliminer l'ambiguïté:

23) ظهرت فجأة **عن يميني** بعد أحد المنعطفات راهبة واقفة وحدها إلى جانب الطريق و نظرها متجه نحوي. (p.112)

*Soudain, **à ma droite,** après un tournant, une Sœur apparut, debout seul au bord de la route et son regard se dirigeait vers moi.*

---

[1] « La relation intrinsèque est une relation binaire R (F, G) entre la cible ('figure') et le site ('ground') qui désigne typiquement une partie du site présentant naturellement un certain type d'orientation (la façade de la maison, la main gauche d'une personne, etc.). » (Costăchescu, 2008b, pp.46)

### 3.1.5. La sémantique de la préposition تحت "sous"

Dans l'énoncé (25), la préposition **تحت** pointe sur la surface cachée de l'entité site.

24) لكنهما ركزا نظرهما بدهشة وخوف على يدي اليُمنى. نظرت بدوري إليها. وجدتها ما تزال ملوثة بالدم، و قد نسي الممرضون غسلها. أسرعت إلى إخفائها **تحت** الغطاء. (p.104)

*Mais ils fixèrent leur regard avec surprise et peur sur ma main droite. Je la regardai à mon tour. Je la trouvai toujours couverte de sang et que les infirmiers ont oublié de la nettoyer. Je me suis précipité pour la cacher **sous** la couverture.*

Dans l'énoncé (26), la cible (رفيقتي وأنا) et le site (صخور) sont repérés l'un par rapport à l'autre selon une direction verticale. Nous remarquons également que le contact entre les entités est possible comme dans l'énoncé (26) mais pas nécessaire comme dans l'énoncé (27).

25) كنا تقدم، رفيقتي وأنا... **فوق** نتوءات الصخور... (p.67)

*Nous étions en train de s'avancer, mon amie et moi, **sur** les roches.*

26) تأخر الوقت وسرنا طويلًا كاميليا وأنا جنبًا إلى جنب في عتمة " شاطئ النخلتين" **تحت** سماء مرصعة بالنجوم. (p.74)

*C'était déjà tard et nous marchâmes, Camélia et moi, l'un contre l'autre dans la nuit de la "plage des deux palmes" **sous** un ciel étoilé.*

Dans certains cas, le pole positif de la relation verticale est exprimé par un verbe qui exprime une direction ascendante de la cible :

27) صعدتُ إلى الطبقة العليا (p.179)

*Je suis monté à l'étage supérieur.*

### 3.1.6. La sémantique des prépositions "أمام" et "وراء" "devant et derrière"

Pour désigner la partie frontale d'un objet, on utilise le substantif « مقدِّمة » (façade). « واجهة » ou « الجزء الأمامي » ou bien (*début de la ligne*).

28) وقعنا على " فندق حايك" ، كما هو مكتوب **على واجهته** بأحرف لاتينية كبيرة عبث بها الزمن. (P.134)

*Nous nous sommes tombés sur l'hôtel Hayek comme il est écrit **sur sa façade** avec des lettres latines majuscules que le temps a changées.*

De même pour وراء et خلف (derrière), pour désigner la partie arrière d'un objet, on utilise les substantifs « مؤخرة », ou bien « الجزء الخلفي » (l'arrière). Les nominalisations أمام (devant), خلف, وراء ne s'appliquent donc qu'aux objets intrinsèquement orientés.

Dans l'énoncé (30), la préposition أمام permet de définir un lieu relativement au « باب المعهد » (*porte de l'institut*), lieu déductible de la position de l'énonciateur. En effet, l'inacceptabilité de l'énoncé (31) revient au fait qu'une porte n'est pas intrinsèquement orientée, elle ne peut être caractérisée par une partie « devant et derrière » "مقدّمة الباب" et "مؤخرة الباب"; cet énoncé décrit l'homme qui se trouve devant la porte. Notons que l'astérisque (*) marque l'agrammaticalité de l'énoncé.

29) كان الرجل يأتي كل يوم من المدينة المجاورة ليبيع الحلوى **أمام** باب المعهد. (p.138)

*L'homme vient tous les jours de la ville voisine pour vendre les friandises **devant** la porte de l'institut.*

30) * كان الرجل يأتي كل يوم من المدينة المجاورة ليبيع الحلوى **في مقدّمة** باب المعهد.

*\*L'homme vient tous les jours de la ville voisine pour vendre les friandises **à l'avant** de la porte de l'institut.*

Au contraire, dans l'énoncé (32), l'ensemble des gens constitue un seul bloc, dans sa globalité et peut être caractérisé par une partie devant et derrière « مقدّمة المشيعين et مؤخرة » et donc le contexte qui permet de lui affecter une orientation, cette orientation permettant d'associer un lieu considéré comme se trouvant « devant ».

31) لم يشأ أن يسير **في مقدمة** المشيعين. (p.37)

*Il ne voulait pas marcher **dans le devant** du convoi.*

Nous pouvons noter que l'expression مقدّمة désigne une partie de l'entité (الجزء الأمامي), tandis que la préposition أمام construit un lieu repère à partir de l'attribution d'une orientation à l'entité.

\* **مقدمة** المنزل كان ضبابيًّا.

*\*Le devant de la maison était brumeux.*

L'inacceptabilité de cet énoncé revient du fait que مقدمة « l'avant » d'un objet n'inclut aucun point de son complément (الجو في الخارج كان ضبابيًّا).

Les adverbiaux مقابل، قبالة، أمام، بوجه، يقابل (*la zone frontale, en face*) introduisent une orientation « en miroir[2] » entre la cible et le site. Dans l'énoncé (33), si le locuteur est assis en face de la peinture de la femme, alors le locuteur est repéré par rapport au lieu « devant » de la peinture, tout en étant orienté vers elle. D'une manière similaire, la peinture est ici orientée (la face colorée de la peinture constitue son devant et donc engendre un lieu « devant »).

32) أمعنّا النظر في مجمل اللوحات، وتوقفت لورا على نحو خاص **أمام** (لوحة) "حاملة المظلة" لأرستيد مايول. (p.93)

*Nous avons bien contemplé toutes les toiles de peinture et Laura s'est arrêtée particulièrement **devant** (la toile) « la porteuse de la parapluie » d'Aristide Mayol.*

### *La direction*

La direction en arabe peut être exprimée par plusieurs verbes tels que : اتجه، تحرك « se diriger, se déplacer » qui expriment le mouvement auxquels s'ajoute l'idée de la direction exprimée par les prépositions surtout نحو « vers » comme l'énoncé (34).

« نحو » est la préposition la plus fréquente qui exprime la direction spatiale. On trouve une deuxième locution prépositionnelle aussi bien utilisée "باتجاه", c'est le cas de l'énoncé (35).

33) واتجهَت **نحو** بلدة مرسى البحرية، فلحقت بها. (p.52)

*Elle s'est dirigée **vers** la ville Marsa marine et je l'ai suivie.*

34) أقفلت عائدًا **في اتجاه** الشاطئ.(p.53)

*Je suis retourné **vers** la plage.*

## 4. Carte sémantique

L'exploitation d'une information spatiale requiert des formalismes de représentation adaptés aux caractéristiques de ce type d'information et compatibles avec le Web sémantique (Berners-Lee et *al,* 2001).
« Une carte sémantique est le produit d'une conceptualisation des relations sémantiques dans les textes. Elle s'exprime par une ontologie linguistique liée à une tâche de fouille textuelle spécifique » (Atanassova, 2012, 100). Selon (Desclés, 2006a), la carte sémantique est un graphe dont les nœuds

---

[2] L'orientation 'en miroir' désigne une position 'face à face' de la cible et du site, par exemple la position de deux personnes, Jean et Paul qui se trouvent face à face (deux personnes qui se parlent, deux boxeurs, les acteurs et les spectateurs, etc.). Comme dans un miroir, la gauche de Jean correspond à la droite de Paul et vice versa. (Costăchescu 2008b)

sont des classes de concepts, et dont les arcs orientés représentent des liaisons de spécifications et de généralisations entre ces classes.

Une carte sémantique est construite à partir d'une étude linguistique profonde ayant pour but de systématiser les marqueurs linguistiques - les prépositions - par lesquels les concepts se réalisent à la surface des textes: l'intérêt de cette carte est d'organiser le domaine spatial en nous laissant guider en premier lieu par les phénomènes linguistiques, observés à partir d'une analyse de corpus, dans un cadre théorique bien précis, sous trois volets: la relation projective, topologique et directionnelle. On donne des exemples sur quelques classes sémantiques qui apparaissent dans la carte :

- نزلت في هذا الفندق exprime une relation topologique d'inclusion de contenace.
- تلك الأسرة الموضوعة تحت أشجار الكينا exprime une relation projective orientationelle verticale.
- نهضنا من النوم متأخرين على زقزقة جوقات العصافير في محيط هذه الشقة exprime une relation projective, une distance qualitative de proximité.
- الآتين من شمال اللوار exprime une relation directionnelle cardinale.

Dans le contexte de notre étude, la carte sémantique liée à la spatialité est une formalisation des connaissances spatiales. La construction de cette carte a été réalisée manuellement. (*Figure 1*)

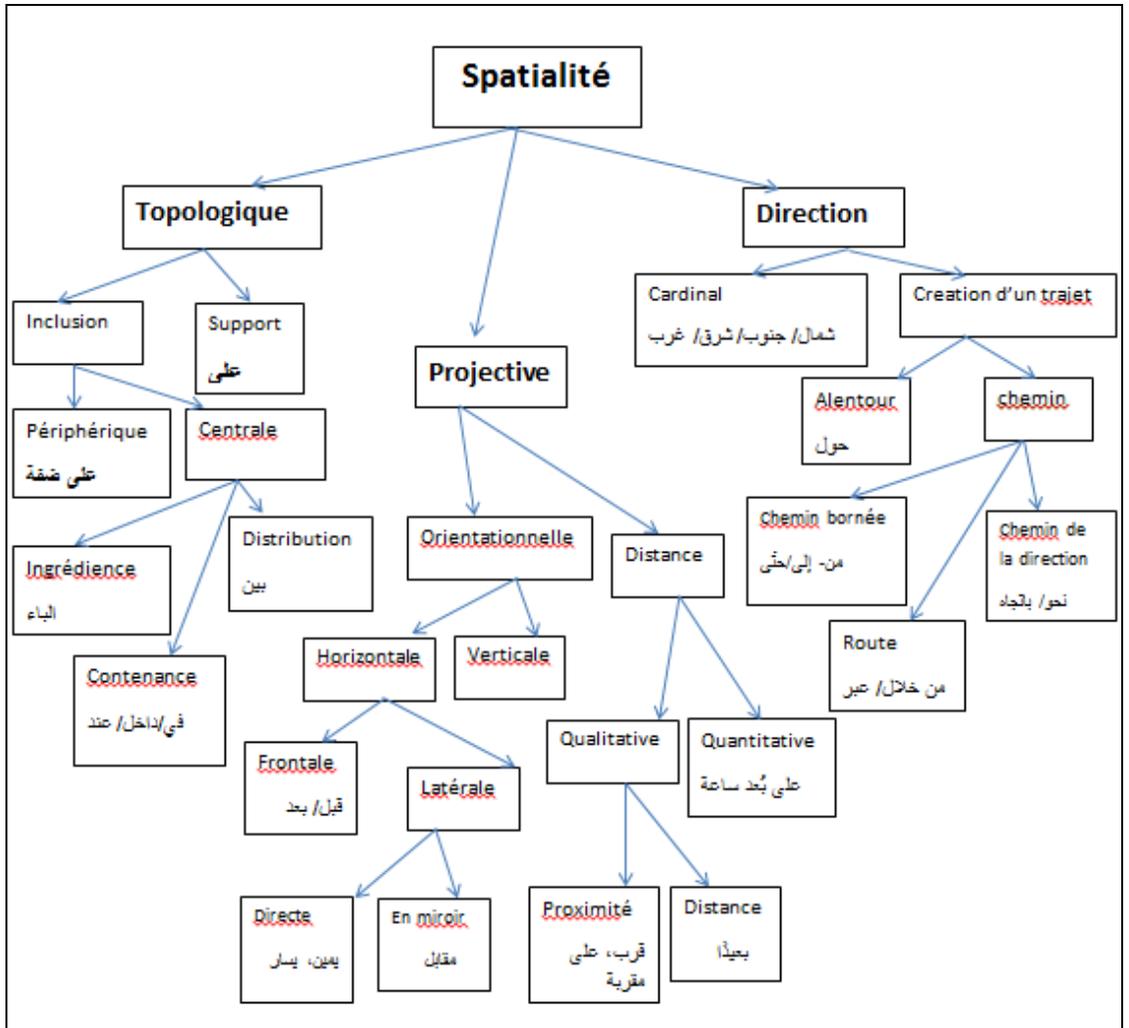

Figure 1 : carte sémantique du domaine spatial

## 5. Reconnaissance automatique des expressions de localisation

Cette phase consiste à mettre en œuvre un système d'annotation sémantique des expressions de localisation. Dans notre approche, nous avons opté pour un système à base d'un ensemble de grammaires locales, implémentés manuellement sous la forme de transducteurs NooJ. La mise en place de règles d'annotation a nécessité une recherche profonde sur certains traits linguistiques propres aux entités spatiales et du domaine spatial en arabe.

Notons que l'annotation se fait juste sur les trois catégories principales (relations topologique, projective et directionnelle). Elle ne produit pas une description complète et précise des relations spatiales.

Le déroulement de cette étape s'effectue en trois temps : la formulation des requêtes, la construction des listes semi-exhaustives, puis la construction des transducteurs afin de les appliquer sur un corpus.

## 5.1. Construction des Règles d'annotation

### 5.1.1. NooJ : une plateforme de développement linguistique

NooJ est un environnement de développement linguistique permettant de construire, de tester et de maintenir des descriptions formalisées à large couverture des langues, sous forme de dictionnaire et de grammaire électroniques dans le but de les appliquer en temps réel à de gros corpus. Cette plateforme présente des fonctionnalités variées du Traitement Automatique des Langues Naturelles: la morphologie flexionnelle et dérivationnelle, les variations terminologiques et orthographiques, le vocabulaire (mots simples, mots composés, expressions figées), la syntaxe et la sémantique, ainsi que comme système de traitement de corpus, et en enseignement des langues.

### 5.1.2. Formulation des requêtes

La première étape de l'étude consiste à rechercher des adverbiaux de lieu tout en prenant en considération la position qu'ils occupent dans la phrase. Puis, la création manuelle d'un module de règles grammaticales : les différentes formes que peut avoir une expression de localisation, et enfin l'implémentation de ce module de règles grammaticales dans NooJ.

Nous optons pour une approche qui repose un système de grammaire implémenté sous forme de transducteurs électroniques NooJ.

Une étape primordiale lors de l'extraction des connaissances consiste à repérer les déclencheurs afin de formuler une requête. Dans notre étude, ces déclencheurs sont des mots présents dans le corpus.

L'étape suivante procède au repérage des expressions locatives à l'aide de patrons :

Les patrons qui repèrent les lieux, les sites : "أمريكا"| "البحيرة"| "اللواريه"|
Les patrons qui repèrent les verbes :"تنساب" | "يقود" | "انتقل" | "تجولت"
Les patrons qui repèrent les prépositions : "إلى"| "على"| "من"| "عن"
Les patrons qui repèrent les entités cibles : "مهرج"| "حبيب"| "كتاب"| "أخي"

Il est nécessaire qu'un déclencheur soit présent dans l'entrée afin d'être en mesure d'extraire l'information présente. A partir d'un déclencheur, un concept est obtenu ce qui nous permet ensuite de sélectionner les règles à appliquer.

Un graphe principal (figure 2) est subdivisé en différents sous-graphes de la façon suivante : des sous-graphes pour les patrons déjà cités, correspondant à la succession des opérations dans les relations spatiales (topologique, directionnelle et projective).

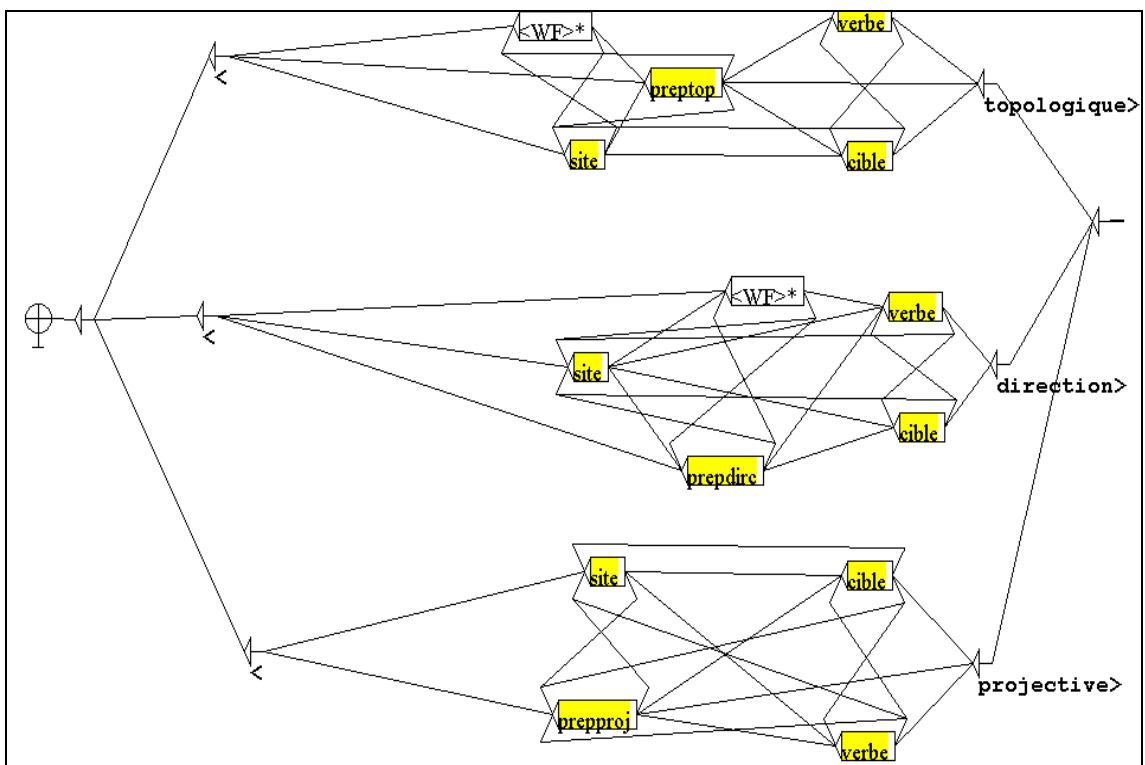

Figure 2: copie d'écran d'un graphe NooJ pour l'annotation des adverbiaux de localisation spatiale

Si les relations syntaxiques correspondent aux règles définies, elles pourront alors être annotées.

### 5.1.3. Premiers résultats

Nous avons appliqué toutes les grammaires construites à un corpus pour vérifier leur validité.

Dans cette section, nous présentons les résultats des évaluations effectuées sur notre corpus textuels pour l'annotation des expressions de localisation. Nous affichons les résultats de l'application de notre approche sur l'annotation des termes sur un sous-ensemble de mots de notre corpus.

## 5.2. Les mesures statistiques

Traditionnellement, l'évaluation de tout système de gestion d'informations repose sur le calcul d'un ensemble de métriques. Ces calculs permettent d'évaluer la proportion des erreurs affichées par le système par rapport au résultat idéal afin de juger la justesse de l'annotation. Les métriques habituellement utilisées sont le Rappel, la Précision et la F-mesure. Pour évaluer le système, nous avons donc comparé les sorties du système avec celles qui ont été produites manuellement.

Sur la base des résultats obtenus par ce traitement manuel et les résultats obtenus par notre système, la précision et le rappel ont été calculés comme indique le tableau 1.

L'évaluation repose sur la comparaison des résultats obtenus automatiquement dans le corpus et celles posées manuellement dans ce même corpus ; des mesures de rappel, de précision et de F-mesure permettent d'objectiver les résultats obtenus.

A l'issue de toutes ces étapes de formalisation, l'annotation automatique des expressions de localisation aboutie à un résultat encourageant. En effet une évaluation effectuée sur notre corpus a permis d'avoir les valeurs suivantes sur la capacité du système à extraire chacun des trois types de relations principales.

|  | **Rappel (R)** | **Précision (P)** | **F-mesure** |
|---|---|---|---|
| **Relation topologique** | 0.81 | 0.77 | 0.79 |
| **Relation projective** | 0.89 | 0.76 | 0.82 |
| **Direction** | 0.83 | 0.75 | 0.79 |

**Tableau 1: Evaluation du système d'annotation des expressions de localisation NooJ**

Bien que spectaculaires, les résultats affichés ne sont pas parfaits. Ces statistiques montrent que, contrairement à ce que l'intuition pourrait laisser penser, le système a des difficultés à déterminer le type d'une relation spatiale.

## 5.3. Analyse des erreurs

Les **bruits** sont principalement liés :
- à la constitution de mes lexiques : le système a extrait l'expression « كاميليا بونار بطاقة من ».
  Cela revient à l'ambiguïté sémantique posée par la préposition مِن. Elle est polysémique.

- à la grammaire qui n'est pas assez ou trop restrictive : "مقيم في ردهة نفسي". La règle d'annotation est correcte, mais cette localisation est abstraite. Or, on cherche juste la localisation concrète, un lieu concret physiquement parlant.
- L'annotation double de quelques expressions: il existe des prépositions qui expriment simultanément plusieurs relations spatiales entre la cible et le site. C'est le cas des prépositions **حول**et **في** **محيط** qui flottent entre deux sens : dans l'espace qui environne (une direction latérale). 2. Dans le voisinage (proximité).
- au marqueur de la négativité non pris en considération par le système, c'est le cas des expressions :
"تمالكت نفسي كي **لا** أقع على الأريكة." "**لم** يحضر هنري مارتي إلى هنا أثناء غيابي".

Les **silences** sont dûs :
- aux expressions trop complexes difficilement modélisables avec les grammaires. Quelques lettres isolées interviennent comme proclitiques de coordination telles que الواو- الفاء. Ils apparaissent généralement accompagnés d'un nom propre de lieu :
"عندما استفقت صباحًا، شعرت برغبة قوية في الابتعاد عن كل ما له علاقة بلورا. عن هذه الشقة، وعن بيت اللواريه، عن باريس، وأورليان، وتور، وكانّ، وأرل، وبروج، و كابورغ، وكل المدن والأنحاء الأخرى التي اعتدنا ارتيادها ".
- la non reconnaissance d'une expression vient d'une déficience du jeu de règles d'annotation et d'une insuffisance de classement sémantique (les *tokens* ne sont pas reconnus parce qu'aucune règle ne les décrits, ou bien les entités ne se trouvent pas dans les listes d'entrée).
- à certains cas ambigus : certaines erreurs proviennent de l'ambiguïté que posent les noms de lieux : il s'agit des noms des lieux étrangers en arabe. La prise en compte des variantes orthographiques des noms propres transcrits en l'absence de conventions pour leurs écritures (notamment pour les noms de lieux) : en arabe, la translittération et la transcription des noms propres étrangers n'obéissent pas à des règles d'écritures. Le mot « Saint » est parfois transcrit « سين جيرمان, », dans d'autres places on écrit (سان لويس), ce qui affecte la précision.

Ces mauvaises délimitations traduisent une déficience au niveau des règles d'annotation qui doivent pouvoir être améliorées.

## 6. Conclusion

Notre travail visait un double objectif : nous avons tenté d'apporter une contribution à la fois théorique, en proposant une analyse linguistique de la spatialité en arabe, et appliquée, en construisant une base de règles informatiques permettant d'extraire automatiquement les phrases relevant de la localisation et de la direction.

Sur le plan de l'analyse linguistique, on a essayé, en se basant toujours sur notre corpus, d'étudier des prépositions spatiales et directionnelles en arabe. En deuxième lieu, une carte sémantique (qui résulte de l'étude des observables) qui propose une organisation originale des modalités dans laquelle *la spatialité* apparaît comme une modalité centrale et organisatrice. Cette organisation sous forme de réseau constitue le point central du travail.

Du côté applicatif, nous avons présenté un module pour les traitements des adverbiaux de localisation spatiale par le biais d'un système à base de règles utilisant des grammaires locales NooJ. En plus d'apporter une contribution théorique, l'étude menée dans le cadre de notre travail nous a permis d'envisager l'écriture de règles informatiques permettant l'annotation automatique des expressions de localisation spatiales et de la direction présents dans les textes. Nous avons ainsi présenté un système de transduction des adverbiaux.

Les problèmes décrits plus hauts sont dus à la difficulté de traiter l'arabe et surtout les prépositions qui portent à la fois la valeur spatiale et temporelle, ainsi, on peut exprimer l'inclusion par plus quatre prépositions, ce qui entraine une ambiguïté au niveau de l'industrie des langues. La méthode préconisée semble être adéquate et affiche des taux de reconnaissance que nous jugeons de bonne qualité. Notre méthode d'annotation montre le caractère indispensable d'une analyse syntaxique dans le repérage de telles informations.

De nombreuses voies restent à explorer dans cette direction. Dans le cadre des applications de TAL, et plus précisément ce qui concerne le domaine de la gestion d'information, notre contribution devra être largement complétée et enrichie d'autres points de vue 1) d'un point de vue pragmatique : étudier la relation spatio-temporelle et faire un système d'annotation des expressions qui marquent la temporalité et celle qui marquent la spatialité. 2) d'un point de vue psycho-cognitive pour étudier la perception et la catégorisation du domaine spatial et 3) la Grammaire Applicative et Cognitive, et plus particulièrement le niveau des représentations sémantico-cognitives des adverbes de localisation.

## 7. Bibliographie